# Rethinking Collapsed Variational Bayes Inference for LDA

**Issei Sato**  SATO@R.DL.ITC.U-TOKYO.AC.JP
The University of Tokyo

**Hiroshi Nakagawa**  N3@DL.ITC.U-TOKYO.AC.JP
The University of Tokyo

## Abstract

We propose a novel interpretation of the collapsed variational Bayes inference with a zero-order Taylor expansion approximation, called CVB0 inference, for latent Dirichlet allocation (LDA). We clarify the properties of the CVB0 inference by using the $\alpha$-divergence. We show that the CVB0 inference is composed of two different divergence projections: $\alpha = 1$ and $-1$. This interpretation will help shed light on CVB0 works.

## 1. Introduction

Latent Dirichlet allocation (LDA) (Blei et al., 2003) is a well-known probabilistic latent variable model. It is used to model the co-occurrence of words by using latent variables called topics where a document is represented as a "bag of words". It has a wide variety of applications in many fields. Originally, the variational Bayes (VB) inference was used for learning LDA. The collapsed variational Bayes (CVB) inference was developed as an alternative deterministic inference for LDA (Teh et al., 2007). The CVB inference is a variational inference improved by marginalizing out parameters as in a collapsed Gibbs sampler (Griffiths & Steyvers, 2004). (Sung et al., 2008) generalized the CVB inference for conjugate-exponential family models, called latent-space variational Bayes (LSVB) inference.

Since the CVB inference requires intractable integrals, Teh et al. (Teh et al., 2007) used a second-order Taylor expansion to perform the integrals. Asuncion et al. (Asuncion et al., 2009; Asuncion, 2010) proposed another approximation that uses only the zero-order information, called the CVB0 inference. The CVB0 inference does not have the drawbacks that other inferences do: VB contains digamma functions which are computationally expensive, while CVB requires the maintenance of variance counts. In contrast, the stochastic nature of the collapsed Gibbs sampler causes it to converge more slowly than the deterministic algorithms. Asuncion et al.'s empirical results suggest that the CVB0 inference learns models that are as good as or better than those learned by the VB and CVB inferences and the collapsed Gibbs sampler in terms of perplexity. Furthermore, as shown in (Asuncion, 2010), when the asymmetric Dirichlet parameters are estimated over document-topic distribution, the predictive performance of the CVB0 inference clearly outperforms that of the CVB inference.

Table 1. Main results: CVB0 is a special case of $\alpha$-divergence projection. CVB0 is interpreted as follows: The ($\alpha = 1$)-divergence is used to estimate $n_{d,t}$, which is the number of times topic $t$ appears in document $d$. The ($\alpha = 1$) divergence is used to estimate $n_{t,v}$, which is the number of times word $v$ is generated from topic $t$. The ($\alpha = -1$) divergence is used to estimate $n_t$, which is the number of times topic $t$ appears in the all documents. "EP" indicates the expectation propagation proposed for an aspect model in (Minka & Lafferty, 2002). In this table, the approximation by Taylor expansion is not assumed with "CVB". "Marginalization" indicates marginalizing out parameters of LDA.

| Inference | Marginalization | $\alpha$-divergence |
|---|---|---|
| VB | NA | $\alpha \to 0$ |
| CVB | ✓ | $\alpha \to 0$ |
| CVB0 | ✓ | $\alpha = 1$ for $n_{d,t}, n_{t,v}$<br>$\alpha = -1$ for $n_t$ |
| EP | NA | $\alpha = 1$ |

We have the question of why CVB0 outperformed CVB, even though the approximation of CVB is more accurate than that of CVB0. In this paper, we propose an interpretation of the CVB0 inference for LDA by using the $\alpha$-divergence. Using the $\alpha$-divergence helps clarify the properties of the CVB0 inference. We also experimentally show the performance of the subspecies of the CVB0 inference, which is derived with the $\alpha$-divergence projection framework. Our analysis of the relationship between existing inference algorithms and $\alpha$-divergence is summarized in Table 1, the meaning of which is revealed in later sections.

The remainder of this paper is organized as follows. Sections 3 and 4 review LDA and the CVB / CVB0 inference for LDA, respectively. Sections 5 and 6 explain $\alpha$-divergence and its local projection, respectively. The key sections 7 and 8 describe local $\alpha$-divergence projection for LDA and its connection to the CVB0 inference. Section 9 introduces other local projections inspired by the CVB0 inference. Section 10 evaluates algorithms in terms of document modeling. Section 11 concludes this paper.

## 2. Preliminaries

Suppose that we have $N$ documents, $V$ vocabularies, and $T$ topics. $\boldsymbol{w} = \{\boldsymbol{w}_d\}_{d=1}^N$ denotes a set of documents and $\boldsymbol{z} = \{\boldsymbol{z}_d\}_{d=1}^N$ is a set of assigned topics. $\theta_{d,t}$ denotes the probability of topic $t$ appearing in document $d$. $\phi_{t,v}$ denotes the probability of word $v$ appearing in topic $t$.

$n_{d,t}(\boldsymbol{z})$ denotes the number of observations of topic $t$ in document $d$. $n_d$ denotes the total number of words in document $d$. $n_{t,v}(\boldsymbol{w}, \boldsymbol{z})$ denotes the number of observations of word $v$ assigned to topic $t$ and $n_{t,\cdot}(\boldsymbol{z}) = \sum_v n_{t,v}(\boldsymbol{w}, \boldsymbol{z})$. For simplicity, we denote them by $n_{d,t}$, $n_{t,v}$ and $n_{t,\cdot}$. The superscript "\d, i" denotes the corresponding variables or counts with $w_{d,i}$ and $z_{d,i}$ excluded, e.g., $\boldsymbol{w}^{\backslash d,i} = \boldsymbol{w} \backslash \{w_{d,i}\}$, $\boldsymbol{z}^{\backslash d,i} = \boldsymbol{z} \backslash \{z_{d,i}\}$, and $n_{t,v}^{\backslash d,i}$ is the number of observations of word $v$ assigned to topic $t$ leaving out $z_{d,i}$.

$\mathbb{E}[x]$ denotes the expectation of $x$ and $\mathbb{V}[x] = \mathbb{E}[x^2] - \mathbb{E}[x]^2$ the variance. Multi$(\cdot)$ denotes the multinomial distribution. Dir$(\cdot)$ denotes the Dirichlet distribution.

## 3. Overview of LDA

The following generative process is assumed with LDA. First, document-topic distribution $\boldsymbol{\theta}_d$ and topic-word distribution $\boldsymbol{\phi}_k$ are generated by

$$\boldsymbol{\theta}_d \sim \text{Dir}(\boldsymbol{\gamma}), \quad \boldsymbol{\phi}_t \sim \text{Dir}(\boldsymbol{\beta}), \quad (1)$$

where $\boldsymbol{\gamma} = (\gamma_1, \cdots, \gamma_T)$ is a $T$-dimensional vector and $\boldsymbol{\beta} = (\beta_1, \cdots, \beta_V)$ is a $V$-dimensional vector.

For each document $d$, generate the $i$-th topic $z_{d,i}$ and word $w_{d,i}$:

$$z_{d,i} \sim \text{Multi}(\boldsymbol{\theta}_d), \quad w_{d,i} \sim \text{Multi}(\boldsymbol{\phi}_{z_{d,i}}). \quad (2)$$

Wallach et al. (Wallach et al., 2009) explored the effects of choosing $\boldsymbol{\gamma}$ and $\boldsymbol{\beta}$ in LDA. They found in Markov chain Monte Carlo (MCMC) simulations that using asymmetric $\boldsymbol{\gamma}$ and symmetric $\boldsymbol{\beta}$ results in better predictive performance for held-out documents. Therefore, we use asymmetric $\boldsymbol{\gamma} = (\gamma_1, \cdots, \gamma_T)$ and symmetric $\boldsymbol{\beta} = (\beta, \cdots, \beta)$.

The assignment probability of topic $t$ to the $i$-th word in document $d$ given $\boldsymbol{w}^{\backslash d,i}, \boldsymbol{z}^{\backslash d,i}, \boldsymbol{\gamma}$ and $\beta$ is

$$\begin{aligned} &p(z_{d,i} = t | w_{d,i} = v, \boldsymbol{w}^{\backslash d,i}, \boldsymbol{z}^{\backslash d,i}, \boldsymbol{\gamma}, \beta) \\ &\propto p(w_{d,i} = v, \boldsymbol{w}^{\backslash d,i}, \boldsymbol{z}^{\backslash d,i}, z_{d,i} = t | \boldsymbol{\gamma}, \beta), \\ &\propto p(w_{d,i} = v | z_{d,i} = t, \boldsymbol{w}^{\backslash d,i}, \boldsymbol{z}^{\backslash d,i} | \beta) p(z_{d,i} = t | \boldsymbol{z}^{\backslash d,i} \boldsymbol{\gamma}), \\ &\propto \frac{n_{t,v}^{\backslash d,i} + \beta}{n_{t,\cdot}^{\backslash d,i} + V\beta} (n_{d,t}^{\backslash d,i} + \gamma_t). \end{aligned} \quad (3)$$

This is used for the collapsed Gibbs sampler.

## 4. CVB/CVB0 inference for LDA

(Teh et al., 2007) proposed the CVB inference to LDA inspired by the collapsed Gibbs sampler and showed that the CVB-LDA outperformed the VB-LDA in terms of perplexity. They only introduced a variational posterior $q(\boldsymbol{z})$ by marginalizing out $\boldsymbol{\theta}$ and $\boldsymbol{\phi}$. The free energy of the CVB-LDA is given by

$$\mathcal{F}_{CVB}[q(\boldsymbol{z})] = \sum_{d=1}^M \sum_{\boldsymbol{z}_d} q(\boldsymbol{z}_d) \log \frac{p(\boldsymbol{w}_d, \boldsymbol{z}_d | \boldsymbol{\gamma}, \beta)}{q(\boldsymbol{z}_d)}. \quad (4)$$

Thus, the updates for $q(\boldsymbol{z})$ are obtained by taking derivatives of $\mathcal{F}_{CVB}[q(\boldsymbol{z})]$ with respect to $\{q(z_{d,i})\}$ and equating to zero:

$$\begin{aligned} &q(z_{d,i} = t) \\ &\propto \exp \mathbb{E}[\log p(w_{d,i} = v, \boldsymbol{w}^{\backslash d,i}, \boldsymbol{z}^{\backslash d,i}, z_{d,i} = t | \boldsymbol{\gamma}, \beta)]_{q(\boldsymbol{z}^{\backslash d,i})}, \\ &\propto \exp \left\{ \mathbb{E}\left[ \log \frac{n_{t,v}^{\backslash d,i} + \beta}{n_{t,\cdot}^{\backslash d,i} + V\beta} (n_{d,t}^{\backslash d,i} + \gamma_t) \right] \right\}, \\ &\propto \frac{\exp \mathbb{E}[\log(n_{t,v}^{\backslash d,i} + \beta)]}{\exp \mathbb{E}[\log(n_{t,\cdot}^{\backslash d,i} + V\beta)]} \exp \mathbb{E}[\log(n_{d,t}^{\backslash d,i} + \gamma_t)]. \quad (5) \end{aligned}$$

This update equation for $q(\boldsymbol{z})$ requires approximations to compute intractable expectation. By using the central limit theorem, the expectation should be closely approximated using Gaussian distributions with means and variances, e.g.,

$$\mathbb{E}[n_{d,t}] = \sum_{i=1}^{n_d} q(z_{d,i} = t), \tag{6}$$

$$\mathbb{V}[n_{d,t}] = \sum_{i=1}^{n_d} q(z_{d,i} = t)(1 - q(z_{d,i} = t)). \tag{7}$$

Moreover, using the second order Taylor expansion, we can approximately calculate

$$q(z_{d,i} = t) \propto \frac{\beta + \mathbb{E}[n_{t,w_{d,i}}^{\setminus d,i}]}{V\beta + \mathbb{E}[n_{t,\cdot}^{\setminus d,i}]}(\gamma_t + \mathbb{E}[n_{d,t}^{\setminus d,i}])$$

$$\exp\left(-\frac{\mathbb{V}[n_{t,w_{d,i}}^{\setminus d,i}]}{2(\beta + \mathbb{E}[n_{t,w_{d,i}}^{\setminus d,i}])^2} + \frac{\mathbb{V}[n_{t,\cdot}^{\setminus d,i}]}{2(V\beta + \mathbb{E}[n_{t,\cdot}^{\setminus d,i}])^2}\right)$$

$$\exp\left(-\frac{\mathbb{V}[n_{d,t}^{\setminus d,i}]}{2(\gamma_t + \mathbb{E}[n_{d,t}^{\setminus d,i}])^2}\right), \tag{8}$$

where the superscript "$\setminus d, i$" denotes subtracting $q(z_{d,i} = t)$ and $q(z_{d,i} = t)(1 - q(z_{d,i} = t))$.

(Asuncion et al., 2009) showed the usefulness of an approximation using only zero-order information, called the CVB0 inference. The update using only zero-order information is given by

$$q(z_{d,i} = t) \propto \frac{\beta + \mathbb{E}[n_{t,w_{d,i}}^{\setminus d,i}]}{V\beta + \sum_v \mathbb{E}[n_{t,v}^{\setminus d,i}]}(\gamma_t + \mathbb{E}[n_{d,t}^{\setminus d,i}]). \tag{9}$$

We derive this CVB0 inference by using $\alpha$-divergence, which enables us to reveal the relationship among other inference algorithms.

## 5. $\alpha$-Divergence

This section reviews $\alpha$-divergence. A readable introduction is provided in (Minka, 2005).

Let our task be to approximate a complex probabilistic distribution $p(\boldsymbol{x})$ where $\boldsymbol{x} = \{x_1, x_2, \cdots, x_n\}$. We approximate $p(\boldsymbol{x})$ as $q(\boldsymbol{x})$, which is a simple probabilistic distribution, such as fully factorized distribution, i.e., $q(\boldsymbol{x}) = \prod_{i=1}^n q(x_i)$. A basic approach to obtaining $q(\boldsymbol{x})$ is to minimize information divergence such as the Kullback-Leibler divergence:

$$KL[p||q] = \int p(\boldsymbol{x}) \log \frac{p(\boldsymbol{x})}{q(\boldsymbol{x})} + \int (q(\boldsymbol{x}) - p(\boldsymbol{x})) d\boldsymbol{x}, \tag{10}$$

where $p(\boldsymbol{x})$ and $q(\boldsymbol{x})$ do not need to be normalized. By using the KL-divergence, the estimation of $q(\boldsymbol{x})$ is defined by the KL-projection of $p(\boldsymbol{x})$ onto a family of $q(\boldsymbol{x})$ as follows:

$$q^*(\boldsymbol{x}) = \underset{q(\boldsymbol{x})}{\operatorname{argmin}}\, KL[p(\boldsymbol{x})||q(\boldsymbol{x})]. \tag{11}$$

$\alpha$-divergence is a generalization of the KL divergence (Amari, 1985; Trottini & Spezzaferri, 2002; Zhu & Rohwer, 1995), indexed by $\alpha \in (-\infty, \infty)$. The $\alpha$ parameter can be used in different ways by different authors. In this paper, we define $\alpha$-divergence by the convention used in (Minka, 2005):

$$D_\alpha[p||q] = \frac{\int \alpha p(\boldsymbol{x}) + (1-\alpha)q(\boldsymbol{x}) - p(\boldsymbol{x})^\alpha q(\boldsymbol{x})^{1-\alpha} d\boldsymbol{x}}{\alpha(1-\alpha)}, \tag{12}$$

where $p(\boldsymbol{x})$ and $q(\boldsymbol{x})$ do not need to be normalized. If $p = q$, $\alpha$-divergence is zero. Some special cases are

$$D_{-1}[p||q] = \frac{1}{2} \int \frac{(q(x) - p(x))^2}{p(x)} dx \tag{13}$$

$$\lim_{\alpha \to 0} D_\alpha[p|q] = KL[q(\boldsymbol{x})||p(\boldsymbol{x})] \tag{14}$$

$$D_{0.5}[p||q] = 2 \int (\sqrt{q(x)} - \sqrt{p(x)})^2 dx \tag{15}$$

$$\lim_{\alpha \to 1} D_\alpha[p|q] = KL[p(\boldsymbol{x})||q(\boldsymbol{x})] \tag{16}$$

$$D_2[p||q] = \frac{1}{2} \int \frac{(p(x) - q(x))^2}{q(x)} dx. \tag{17}$$

The case $\alpha = 0.5$ is known as the Hellinger distance, and $\alpha = 2$ is the $\chi^2$ distance. Since $\alpha = -1$ swaps the position of $p$ and $q$ of the $\chi^2$ distance, we call the case $\alpha = -1$ "the inverse $\chi^2$ distance", which is the key divergence in this paper.

## 6. Local $\alpha$-divergence projection

In this section, we introduce a local divergence projection-based inference.

Suppose that the approximate distribution $q(\boldsymbol{x})$ is fully factorized. We derive the update $q(x_i)$ minimizing $\alpha$-divergence as follows. Taking derivatives of $\alpha$-divergence (12) with respect to $q(x_i)$ and equating them to zero, we obtain the following fixed point iteration equations:

$$q(x_i) \propto \mathbb{E}\left[\left(\frac{p(\boldsymbol{x})}{q(\boldsymbol{x}^{\setminus i})}\right)^\alpha\right]_{q(\boldsymbol{x}^{\setminus i})}^{\frac{1}{\alpha}} \tag{18}$$

In many cases, this update is intractable and thus we introduce an approximation for Eq. (18).

Since Eq. (18) is

$$q(x_i) \propto \mathbb{E}\left[\left(p(x_i|\boldsymbol{x}^{\setminus i})\frac{p(\boldsymbol{x}^{\setminus i})}{q(\boldsymbol{x}^{\setminus i})}\right)^\alpha\right]_{q(\boldsymbol{x}^{\setminus i})}^{\frac{1}{\alpha}}, \tag{19}$$

we replace $p(\boldsymbol{x}^{\setminus i})$ with $q(\boldsymbol{x}^{\setminus i})$, obtaining

$$q(x_i) \propto \mathbb{E}\left[p(x_i|\boldsymbol{x}^{\setminus i})^\alpha\right]_{q(\boldsymbol{x}^{\setminus i})}^{\frac{1}{\alpha}}. \tag{20}$$

In the case $\alpha = 1$, the update (20) is similar to belief propagation, and the factorized neighbors algorithm (Rosen-Zvi et al., 2005).

The update (20) means that it locally minimize $\alpha$-divergence, i.e., for each $i$,

$$q^*(x_i) = \operatorname*{argmin}_{q(x_i)} D_\alpha[p(x_i|\boldsymbol{x}^{\setminus i})q(\boldsymbol{x}^{\setminus i})\|q(\boldsymbol{x})]. \quad (21)$$

In the case $\alpha = 1$, i.e., KL divergence, this local projection-based inference is equal to the EP algorithm. We describe the connection of this $\alpha$-divergence projection with the CVB0 inference in the next section.

## 7. CVB0 as $\alpha$-divergence projection

In this section, we derive the CVB0 inference by using the local $\alpha$-divergence projection. First, we describe how the case $\alpha = 1$, i.e. EP, cannot be applied for the collapsed LDA. Second, we derive a divergence projection applicable to the collapsed LDA and explain the relationship between this projection and the CVB0 inference.

We apply Eq. (21) with $\alpha = 1$(EP) to the collapsed LDA. For each $z_{d,i}$, we perform

$$q^*(z_{d,i}) = \operatorname*{argmin}_{q(z_{d,i})} KL[p(z_{d,i}|\boldsymbol{w}, \boldsymbol{z}^{\setminus d,i})q(\boldsymbol{z}^{\setminus d,i})\|q(\boldsymbol{z})]. \quad (22)$$

The update for $q(z_{d,i})$ is

$$q(z_{d,i} = t) \propto \mathbb{E}\left[p(z_{d,i} = t|w_{d,i} = v, \boldsymbol{w}^{\setminus d,i}, \boldsymbol{z}^{\setminus d,i})\right]_{q(\boldsymbol{z}^{\setminus d,i})},$$

$$\propto \mathbb{E}\left[(n_{d,t}^{\setminus d,i} + \gamma_t)\frac{n_{t,v}^{\setminus d,i} + \beta}{n_{t,\cdot}^{\setminus d,i} + V\beta}\right]_{q(\boldsymbol{z}^{\setminus d,i})}. \quad (23)$$

The problem is that we cannot analytically execute this expectation. (Asuncion, 2010) derived Eq.(23) in a different way where he changed the CVB free energy by moving the logarithm out of the expectations, and pointed out the relationship between Eq.(23) and the CVB0 inference, which inspired this work. However, the intractable expectation in Eq.(23) was not executed. This intractability makes interpreting the CVB0 inference difficult.

Here, we derive another approach by using the $\alpha$-divergence projection. The key idea is to construct $q(z_{d,i})$ by using the novel three parameters.

We define $q(z_{d,i})$ as follows:

$$q(z_{d,i} = t) \propto a(z_{d,i})b(z_{d,i})c(z_{d,i}) \quad (24)$$
$$a(z_{d,i} = t) = \tilde{n}_{d,t}^{\setminus d,i} + \gamma_t, \quad (25)$$
$$b(z_{d,i} = t) = \tilde{n}_{t,v}^{\setminus d,i} + \beta, \quad (26)$$
$$c(z_{d,i} = t) = \frac{1}{\tilde{n}_{t,\cdot}^{\setminus d,i} + V\beta}, \quad (27)$$

where we do not assume that $\tilde{n}_{d,i}^{\setminus d,i}$, $\tilde{n}_{t,v}^{\setminus d,i}$ and $\tilde{n}_{t,\cdot}^{\setminus d,i}$ are expected counts, i.e., these are parameters of $q(z_{d,i})$.

We also define

$$q^{\setminus a}(z_{d,i}) = b(z_{d,i})c(z_{d,i}), \quad (28)$$
$$q^{\setminus b}(z_{d,i}) = a(z_{d,i})c(z_{d,i}) \quad (29)$$
$$q^{\setminus c}(z_{d,i}) = a(z_{d,i})b(z_{d,i}). \quad (30)$$

Since our definition of $\alpha$-divergence does not require normalization of the probabilistic distribution, we can introduce the following local projection:

$$a^*(z_{d,i} = t) =$$
$$\operatorname*{argmin}_{a(z_{d,i})} D_\alpha[(n_{d,t}^{\setminus d,i} + \gamma_t)q^{\setminus a_{d,i}}(\boldsymbol{z})\|a(z_{d,i})q^{\setminus a_{d,i}}(\boldsymbol{z})], \quad (31)$$

where $q^{\setminus a_{d,i}}(\boldsymbol{z}) = q^{\setminus a}(z_{d,i})q(\boldsymbol{z}^{\setminus d,i})$. Solving the above optimization (see Appendix A), we obtain

$$a^*(z_{d,i} = t) = \mathbb{E}\left[(n_{d,t}^{\setminus d,i} + \gamma_t)^\alpha\right]_{q(\boldsymbol{z}^{\setminus d,i})}^{\frac{1}{\alpha}}. \quad (32)$$

As in $a(z_{d,i})$, we obtain $b^*(z_{d,i})$ and $c^*(z_{d,i})$ by locally minimizing the $\alpha$-divergence:

$$b^*(z_{d,i} = t) =$$
$$\operatorname*{argmin}_{b(z_{d,i})} D_\alpha[(n_{t,v}^{\setminus d,i} + \beta)q^{\setminus b_{d,i}}(\boldsymbol{z})\|b(z_{d,i})q^{\setminus b_{d,i}}(\boldsymbol{z})], \quad (33)$$

$$c^*(z_{d,i} = t) =$$
$$\operatorname*{argmin}_{c(z_{d,i})} D_\alpha[\frac{1}{(n_{t,\cdot}^{\setminus d,i} + V\beta)}q^{\setminus c_{d,i}}(\boldsymbol{z})\|c(z_{d,i})q^{\setminus c_{d,i}}(\boldsymbol{z})]. \quad (34)$$

Thus, we have

$$b^*(z_{d,i} = t) = \mathbb{E}\left[(n_{t,v}^{\setminus d,i} + \beta)^\alpha\right]_{q(\boldsymbol{z}^{\setminus d,i})}^{\frac{1}{\alpha}}, \quad (35)$$

$$c^*(z_{d,i} = t) = \mathbb{E}\left[\left(\frac{1}{n_{t,\cdot}^{\setminus d,i} + V\beta}\right)^\alpha\right]_{q(\boldsymbol{z}^{\setminus d,i})}^{\frac{1}{\alpha}}. \quad (36)$$

When we use $\alpha$-divergence projection with $\alpha = 1$ for estimating $a(z_{d,i})$ and $b(z_{d,i})$, we have

$$a^{(\alpha=1)}(z_{d,i}=t) = \mathbb{E}\left[n_{d,t}^{\backslash d,i} + \gamma_t\right]_{q(\mathbf{z}^{\backslash d,i})} = \mathbb{E}[n_{d,t}^{\backslash d,i}] + \gamma_t, \tag{37}$$

$$b^{(\alpha=1)}(z_{d,i}=t) = \mathbb{E}\left[n_{t,v}^{\backslash d,i} + \beta\right]_{q(\mathbf{z}^{\backslash d,i})} = \mathbb{E}[n_{t,v}^{\backslash d,i}] + \beta. \tag{38}$$

When we use $\alpha$-divergence projection with $\alpha = -1$ for estimating $c(z_{d,i})$, we have

$$\begin{aligned}c^{(\alpha=-1)}(z_{d,i}=t) &= \mathbb{E}\left[\left(\frac{1}{n_{t,\cdot}^{\backslash d,i}+V\beta}\right)^{-1}\right]^{-1}_{q(\mathbf{z}^{\backslash d,i})},\\ &= \mathbb{E}\left[n_{t,\cdot}^{\backslash d,i}+V\beta\right]^{-1}_{q(\mathbf{z}^{\backslash d,i})},\\ &= \frac{1}{\mathbb{E}[n_{t,\cdot}^{\backslash d,i}]+V\beta}. \end{aligned} \tag{39}$$

Therefore, we have the following update for $q(z_{d,i})$

$$\begin{aligned} q(z_{d,i}=t) &\propto a^{(\alpha=1)}(z_{d,i})b^{(\alpha=1)}(z_{d,i})c^{(\alpha=-1)}(z_{d,i}),\\ &= (\mathbb{E}[n_{d,t}^{\backslash d,i}]+\gamma)\frac{\mathbb{E}[n_{t,v}^{\backslash d,i}]+\beta}{\mathbb{E}[n_{t,\cdot}^{\backslash d,i}]+V\beta}. \end{aligned} \tag{40}$$

Although the updates are performed in order, i.e., update $a^*$ given $b$ and $c$, $b^*$ given $a^*$ and $c$, and $c^*$ given $a^*$ and $b^*$, this update is equal to the CVB0 update in Eq.(9).

## 8. Discussion

In this section, we explain why the CVB0 inference outperforms the CVB inference. To sum up this discussion, in the CVB0 inference, the "zero-forcing effect" works only with the $n_{t,\cdot}$ estimation, while in the CVB inference it works with the $q(z)$ estimation.

The previous section showed that the CVB0 inference is composed of the three projections with a mixture of $\alpha = 1$ and $\alpha = -1$:

$$D_1 = KL[(n_{d,t}^{\backslash d,i}(\mathbf{z})+\gamma_t)q^{\backslash a_{d,i}}(\mathbf{z})||q(\mathbf{z})], \tag{41}$$

$$D_1 = KL[(n_{t,v}^{\backslash d,i}(\mathbf{z})+\beta)q^{\backslash b_{d,i}}(\mathbf{z})||q(\mathbf{z})], \tag{42}$$

$$D_{-1}\left[\frac{1}{(n_{t,\cdot}^{\backslash d,i}(\mathbf{z})+V\beta)}q^{\backslash c_{d,i}}(\mathbf{z})||q(\mathbf{z})\right]. \tag{43}$$

This projection-based update with a different divergence measure reveals the properties of the CVB0 inference. Ideally, we use the $(\alpha=1)$-divergence projection, i.e., $D_1[p|q] = KL[p||q]$, but the integrals $\mathbb{E}[\frac{1}{n_{t,\cdot}^{\backslash d,i}+V\beta}]$ are not easy to evaluate. Instead, we use the inverse $\chi^2$ divergence $D_{-1}[p||q]$ for estimating $c(z_{d,i})$.

$D_{-1}[p||q] = \frac{1}{2}\int\frac{(q(x)-p(x))^2}{p(x)}dx$ is known as a zero-forcing divergence (Minka, 2005) which emphasizes $q$ to be small when $p$ being small, i.e., $p(x) = 0$ forces $q(x) = 0$, which means that it avoids "false positive". In our case (43), the zero-forcing effect on the $n_{t,\cdot}$ estimation means that the emphasis in the estimation is on high-frequency topics or low-frequency topics tend to be estimated as zero in an entire corpus. We think that affecting $n_{t,\cdot}^{\backslash d,i}$ matters much less than affecting $n_{d,t}^{\backslash d,i}$ and $n_{t,v}^{\backslash d,i}$ throughout a whole corpus in LDA. We explain the zero-forcing effect of CVB0 in more detail in the next section.

Returning to Eq.(20), i.e., $q(x_i) \propto \mathbb{E}\left[p(x_i|\mathbf{x}^{\backslash i})^\alpha\right]^{\frac{1}{\alpha}}_{q(\mathbf{x}^{\backslash i})}$, we describes the relationship between the CVB inference and $\alpha$-divergence projection. First, we introduce the following theorem:

**Theorem 1 (Liapunov's inequality)** *If $x$ is a non-negative random variable, and we have two real numbers $\alpha_2 > \alpha_1$, then*

$$\mathbb{E}[x^{\alpha_2}]^{\frac{1}{\alpha_2}} \geq \mathbb{E}[x^{\alpha_1}]^{\frac{1}{\alpha_1}}. \tag{44}$$

*and*

$$\lim_{\alpha\to 0}\mathbb{E}[x^\alpha]^{\frac{1}{\alpha}} = \exp\mathbb{E}[\log(x)]. \tag{45}$$

Using Eq.(20) and Theorem 1, we obtain

$$q(x_i) \propto \lim_{\alpha\to 0}\mathbb{E}\left[p(x_i|\mathbf{x}^{\backslash i})^\alpha\right]^{\frac{1}{\alpha}}_{q(\mathbf{x}^{\backslash i})} = \exp(\mathbb{E}[\log p(x_i|\mathbf{x}^{\backslash i})]) \tag{46}$$

This is the variational inference minimizing $KL[q||p]$.

In LDA, we have

$$\begin{aligned} q(z_{d,i}=t) &\propto \lim_{\alpha\to 0}\mathbb{E}\left[p(z_{i,d}|w_{d,i}=v,\mathbf{w}^{\backslash d,i},\mathbf{z}^{\backslash d,i})^\alpha\right]^{\frac{1}{\alpha}}_{q(\mathbf{z}^{\backslash d,i})}\\ &= \exp(\mathbb{E}[\log p(z_{d,i}|w_{d,i}=v,\mathbf{w}^{\backslash d,i},\mathbf{z}^{\backslash d,i})])\\ &\propto \exp\mathbb{E}\left[\log\frac{n_{d,t}^{\backslash d,i}+\gamma_t}{n_{d,\cdot}^{\backslash d,i}+\sum_t\gamma_t}\frac{n_{t,v}^{\backslash d,i}+\beta}{n_{t,\cdot}^{\backslash d,i}+V\beta}\right]_{q(\mathbf{z}^{\backslash d,i})},\\ &\propto \exp(\mathbb{E}[\log(n_{d,t}^{\backslash d,i}+\gamma_t)])\frac{\exp(\mathbb{E}[\log(n_{t,v}^{\backslash d,i}+\beta)])}{\exp(\mathbb{E}[\log(n_{t,\cdot}^{\backslash d,i}+V\beta)])} \end{aligned} \tag{47}$$

The update Eq.(47) is the same update as the CVB inference in Eq.(5). $(\alpha\to 0)$-divergence is also known to induce the zero-forcing effect.

## 9. Subspecies inspired by CVB0

In this section, we consider other projection-based algorithms that help clarify the property of the zero-forcing effect in CVB0.

### 9.1. CVB with $(\alpha = 1)$-divergence

From our view point, the CVB0 inference is composed of two different-type divergence projections: $\alpha = 1, -1$. We consider using only $\alpha = 1$ for the projections. To do this, we have to calculate the expectation given by

$$c^{(\alpha=1)}(z_{d,i}=t) = \mathbb{E}\left[\frac{1}{n_{t,\cdot}^{\backslash d,i} + V\beta}\right]_{q(\mathbf{z}^{\backslash d,i})}. \quad (48)$$

Since we cannot derive the analytical solution for this expectation, we propose two approximation methods. The first is a stochastic approximation called sample averaging given by

$$\tilde{c}^{(\alpha=1)}(z_{d,i}=t) = \frac{1}{S}\sum_{s=1}^{S}\frac{1}{n_{t,\cdot}^{\backslash d,i}(\mathbf{z}^{(s)}) + V\beta}, \quad (49)$$

where $S$ denotes the number of samples and $\mathbf{z}^{(s)}$ is the $s$-th samples generated from $q(\mathbf{z})$. This method is accurate but not practical when $S$ takes a large value. We use this approximation to investigate the accuracy of the next approximation.

The second is a deterministic approximation that uses the same approximation of CVB with the second-order Taylor expansion and Gaussian approximation given by

$$\hat{c}^{(\alpha=1)}(z_{d,i}=t) = \frac{1}{\mathbb{E}[n_{t,\cdot}^{\backslash d,i}] + V\beta} + \frac{\mathbb{V}[n_{t,\cdot}^{\backslash d,i}]}{(\mathbb{E}[n_{t,\cdot}^{\backslash d,i}] + V\beta)^3}. \quad (50)$$

As shown in the experiments (Sec.10), we find that the second term of Eq.(50) is vanishingly small. $\hat{c}^{(\alpha=1)}$ in Eq.(50) is calculated as $\frac{1}{\mathbb{E}[n_{t,\cdot}^{\backslash d,i}] + V\beta}\left(1 + \frac{\mathbb{V}[n_{t,\cdot}^{\backslash d,i}]}{(\mathbb{E}[n_{t,\cdot}^{\backslash d,i}] + V\beta)^2}\right)$. We find $\frac{\mathbb{V}[n_{t,\cdot}^{\backslash d,i}]}{(\mathbb{E}[n_{t,\cdot}^{\backslash d,i}] + V\beta)^2} = O(1/n)$ in many cases where $n$ denotes the number of all words (tokens). For example, the variance takes the largest value when $q(z_{d,i} = t) = 1/2$ for all $d$ and $i$. In this case, $\mathbb{E}[n_t] = n/2$ and $\mathbb{V}[n_t] = n(1 - 1/2)/2 = n/4$. Therefore, we consider $c^{(\alpha=-1)}$ is similar to $c^{(\alpha=1)}$, which means that CVB0 is rarely affected by the zero-forcing effect.

### 9.2. Type-base CVB0 Inference

We derive a type-based inference as an application of our framework. In a type-based inference, we only estimate the probabilistic distribution for each type in a document not each token; this is beneficial for computation cost and memory usage.

We exclude all counts of word $v$ from document $d$, denoted by superscript "$\backslash d, v$". The appearance probability of word $v$ given $\mathbf{w}_d^{\backslash d,v}$ and $\mathbf{z}_d^{\backslash d,v}$ is

$$p(w_{d,*} = v|\mathbf{w}^{\backslash d,v}, \mathbf{z}^{\backslash d,v}) = \sum_{t=1}^{T}\frac{n_{d,t}^{\backslash d,v} + \gamma_t}{n_{d,\cdot}^{\backslash d,v} + \sum_t \gamma_t}\frac{n_{t,v}^{\backslash d,v} + \beta}{n_{t,\cdot}^{\backslash d,v} + V\beta} \quad (51)$$

Moreover, we have

$$p(z_{d,v} = t|\mathbf{w}^{\backslash d,v}) \propto$$
$$\mathbb{E}\left[\frac{n_{d,t}^{\backslash d,v} + \gamma_t}{n_{d,\cdot}^{\backslash d,v} + \sum_t \gamma_t}\frac{n_{t,v}^{\backslash d,v} + \beta}{n_{t,\cdot}^{\backslash d,v} + V\beta}\right]_{p(\mathbf{z}^{\backslash d,v}|\mathbf{w}^{\backslash d,v})} \quad (52)$$

Here, we consider obtaining an approximation distribution $q(z_{d,v})$. Instead of $z_{d,i}$, we define $q(\mathbf{z})$ factorized by using $q(z_{d,v})$, i.e., $q(z_{d,i}) = \sum_{v=1}^{V} q(z_{d,v})\delta(w_{d,i} = v)$ and $q(\mathbf{z}) = \prod_{v=1}^{V} q(z_{d,v})^{n_{d,v}}$.

The update of $q(z_{d,v})$ is obtained by

$$q(z_{d,v} = t) \propto \mathbb{E}\left[(n_{d,t}^{\backslash d,v} + \gamma_t)\frac{n_{t,v}^{\backslash d,v} + \beta}{n_{t,\cdot}^{\backslash d,v} + V\beta}\right]_{q(\mathbf{z}^{\backslash d,v})}, \quad (53)$$

which is derived by minimizing the $\alpha$-divergence as in $q(z_{d,i})$.

Using the local $\alpha$-divergence projection with $\alpha = 1$ for $n_{d,t}^{\backslash d,v} + \gamma_t$ and $n_{t,v}^{\backslash d,v} + \beta$, and $\alpha = -1$ for $\frac{1}{n_{t,\cdot}^{\backslash d,v} + V\beta}$, we have

$$q(z_{d,v} = t) \propto (\mathbb{E}[n_{d,t}^{\backslash d,v}] + \gamma)\frac{\mathbb{E}[n_{t,v}^{\backslash d,v}] + \beta}{\mathbb{E}[n_{t,\cdot}^{\backslash d,v}] + V\beta}. \quad (54)$$

We call this update the type-based CVB0 (TCVB0) inference.

## 10. Experiments

We compared CVB0 with its subspecies on document modeling in terms of perplexity to investigate the effect of $\alpha = -1$. All results are averaged values from five experimental runs with random initialization. We set the number of iterations to 100 for each inference.

We use a fixed point equation for updating $\boldsymbol{\gamma}$ introduced in (Minka, 2000). We set $\beta = 0.01$ because

Rethinking Collapsed Variational Bayes Inference for LDA(Asuncion et al., 2009) showed that CVB0-LDA with $\beta = 0.01$ worked well when compared with other settings ($\beta = 0.01$ was also used in (Griffiths & Steyvers, 2004)).

In this section's figures, "CVB" indicates the second order approximation of the CVB inference.

"CVB1s" indicates the stochastic approximation in Eq.(49) with $S = 50$. "CVB1d" indicates the deterministic approximation in Eq.(50).

We used four sets of text data with different properties. The first was 'NIPS corpus (NIPS)" from which the number of documents was $N = 1,500$ and the vocabulary size was $V = 12,245$. The second was "The Wall Street Journal (WSJ)" from which we randomly chose $N = 5,000$ ($V = 38,272$) documents. The third was "Enron email corpus (Enron)" from which we randomly chose $N = 5,000$ ($V = 14,758$) documents. The fourth was "20 news group corpus (20ng)" from which we randomly chose $N = 5,000$ ($V = 13,176$). Stop words were eliminated.

The comparison metric we used for document modeling was the perplexity used by (Teh et al., 2007; 2008) that indicates the prediction performance for held-out words. We randomly split the words in a document into training words $\boldsymbol{w}_d^{train}$ (80%) and test words $\boldsymbol{w}_d^{test}$ (20%).

Figure 1 shows the experimental results. The bar graph indicates the results for test set perplexity in terms of ($T = 40, 80, 120$) in each corpus. CVB0, CVB1s, CVB1d and TCVB0 outperformed CVB in terms of perplexity. Although we compared VB with others, we eliminated the VB results to clarify the differences of inference algorithms because CVB outperformed VB and the VB results change the scale of a bar-graph in some corpora.

The performances of CVB1s and CVB1d were similar to that of CVB0. Since the results of CVB1d were similar to those of CVB1s, the approximation used in CVB1d seemed to be accurate. When we analyzed $\frac{\mathbb{V}[n_{t,\cdot}]}{(\mathbb{E}[n_{t,\cdot}] + V\beta)^2}$ in Sec.9.1, the maximum value in all corpus when $T = 120$ was about $3.17e^{-4}$, which is negligible compared with 1. Therefore, as discussed in Sec.9.1, CVB0 was not affected by the zero-forcing effect. We believe this is the reason CVB0 worked better than CVB. Moreover, the performance of TCVB0 was similar to that of CVB0. Consequently, the TCVB0 inference was practical.

## 11. Conclusion

In this paper, we reviewed existing inference algorithms of LDA in terms of the $\alpha$-divergence projection. We showed that the CVB0 inference is composed of ($\alpha = 1, -1$)- divergence projections and that $\alpha = -1$ is similar to $\alpha = 1$ in LDA, which means that CVB0 is not affected by the zero-forcing effect in LDA. Combining the marginalization of parameters and the heterogeneous $\alpha$-divergence projection is useful because it is easy to apply to other topic models learned by the collapsed Gibbs sampler. Future work is to develop an online-update extension, such as that by (Hoffman et al., 2010; Sato et al., 2010; Wang et al., 2011). From the relationship between EP and assumed density filtering, we can extend the local $\alpha$-divergence projection into an online algorithm, which leads to the online CVB0 inference. A convergence analysis is also important remaining work.

## A. Derivation for Eq.(32)

Taking derivatives of

$$D_\alpha[(n_{d,t}^{\backslash d,i} + \gamma_t) q^{\backslash a_{d,i}}(\boldsymbol{z}) || a(z_{d,i}) q^{\backslash a_{d,i}}(\boldsymbol{z})],$$

with respect to $a(z_{d,i})$ and equating them to zero,

$$0 = \sum_{\boldsymbol{z}^{\backslash d,i}} q^{\backslash a_{d,i}}(\boldsymbol{z}) - a(z_{d,i})^{-\alpha} \sum_{\boldsymbol{z}^{\backslash d,i}} (n_{d,t}^{\backslash d,i} + \gamma_t)^\alpha q^{\backslash a_{d,i}}(\boldsymbol{z}),$$

and we obtain the following fixed point iteration equations:

$$\begin{aligned} a(z_{d,i}) &= \left[\frac{\sum_{\boldsymbol{z}^{\backslash d,i}} (n_{d,t}^{\backslash d,i} + \gamma_t)^\alpha q^{\backslash a_{d,i}}(\boldsymbol{z})}{\sum_{\boldsymbol{z}^{\backslash d,i}} q^{\backslash a_{d,i}}(\boldsymbol{z})}\right]^{\frac{1}{\alpha}}, \\ &= \left[\frac{\sum_{\boldsymbol{z}^{\backslash d,i}} (n_{d,t}^{\backslash d,i} + \gamma_t)^\alpha b(z_{d,i}) c(z_{d,i}) q(\boldsymbol{z}^{\backslash d,i})}{\sum_{\boldsymbol{z}^{\backslash d,i}} b(z_{d,i}) c(z_{d,i}) q(\boldsymbol{z}^{\backslash d,i})}\right]^{\frac{1}{\alpha}}, \\ &= \left[\frac{b(z_{d,i}) c(z_{d,i}) \sum_{\boldsymbol{z}^{\backslash d,i}} (n_{d,t}^{\backslash d,i} + \gamma_t)^\alpha q(\boldsymbol{z}^{\backslash d,i})}{b(z_{d,i}) c(z_{d,i}) \sum_{\boldsymbol{z}^{\backslash d,i}} q(\boldsymbol{z}^{\backslash d,i})}\right]^{\frac{1}{\alpha}}. \end{aligned} \quad (55)$$

Since $\sum_{\boldsymbol{z}^{\backslash d,i}} q(\boldsymbol{z}^{\backslash d,i}) = 1$, we have

$$a(z_{d,i}) = \left[\sum_{\boldsymbol{z}^{\backslash d,i}} (n_{d,t}^{\backslash d,i} + \gamma_t)^\alpha q(\boldsymbol{z}^{\backslash d,i})\right]^{\frac{1}{\alpha}}. \quad (56)$$

## References

Amari, S. *Differential-Geometrical Methods in Statistic*. Springer, New York, 1985.

Asuncion, A. Approximate mean field for dirichlet-based models. In *Topic Models Workshop, ICML*. 2010.

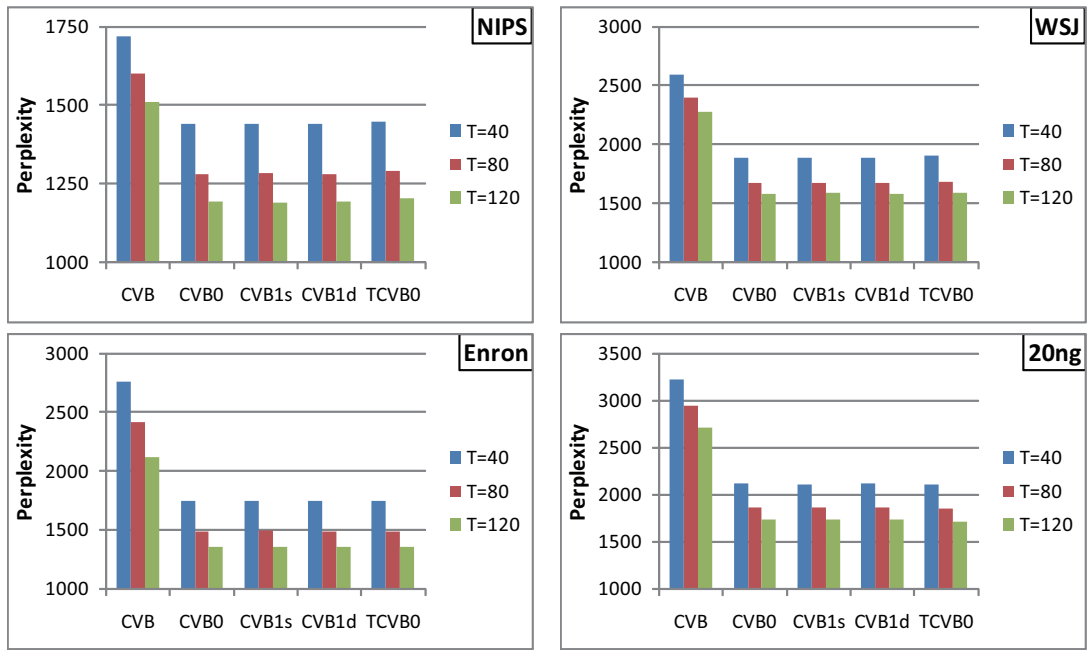

Figure 1. Experiment results for document modeling in four datasets. $T$ denotes the number of topics. Lower perplexity indicates better performance.